\documentclass[letterpaper, 10 pt, conference]{ieeeconf}  % Comment this line out if you need a4paper

\IEEEoverridecommandlockouts                              % This command is only needed if 
\usepackage{graphics} % for pdf, bitmapped graphics files
\usepackage{epsfig} % for postscript graphics files
\usepackage{amsmath} % assumes amsmath package installed
\usepackage{amssymb}  % assumes amsmath package installed
\usepackage{algorithm}
\usepackage{algpseudocode}
\usepackage{amssymb}
\usepackage{amsmath}
\usepackage{commath}
\usepackage{mathtools}

\usepackage{hyperref}
\usepackage{import}
\usepackage{paralist}
\usepackage{gensymb}
\usepackage{dsfont}
\usepackage{siunitx}
\usepackage[bottom]{footmisc}
\usepackage{framed}
\usepackage{balance} 
\usepackage{caption}
\usepackage[skins]{tcolorbox}  
\usepackage{subcaption}
\usepackage{bm}
\usepackage{cite} % [noadjust]
\usepackage{balance}

\hypersetup{
    colorlinks=true,
    linkcolor=black,
    citecolor=black,
    filecolor=black,
    urlcolor=black,
}
\DeclareCaptionFont{mysize}{\fontsize{8}{10}\selectfont}
\usepackage{soul}
\usepackage{hyperref}
\usepackage[percent]{overpic}
\usepackage{xcolor}
\newcommand{\insertWebPageLink}{\url{https://miladshafiee.github.io/puppeteer-and-marionette/}}

\usepackage[subtle,tracking=normal]{savetrees} 

\title{\Large \bf
\textit{Puppeteer and Marionette:}  Learning  Anticipatory Quadrupedal Locomotion Based on  Interactions of a Central Pattern Generator and Supraspinal Drive
}

    \author{Milad Shafiee$^{1}$, Guillaume Bellegarda$^{1}$, and Auke Ijspeert$^{1}$ % <-this % stops a space
\thanks{$^{1}$ This research is supported by the Swiss National Science Foundation
(SNSF) as part of project No.197237. The authors are with the BioRobotics Laboratory, Ecole Polytechnique Federale de Lausanne (EPFL). {\tt\footnotesize (e-mail: firstname.lastname@epfl.ch)}}% <-this % stops a space
}

\begin{document}
\bstctlcite{MyBSTcontrol}
\maketitle
\thispagestyle{empty}
\pagestyle{empty}

%%%%%%%%%%%%%%%%%%%%%%%%%%%%%%%%%%%%%%%%%%%%%%%%%%%%%%%%%%%%%%%%%%%%%%%%%%%%%%%%
\begin{abstract}
Quadruped animal locomotion emerges from the interactions between the spinal central pattern generator (CPG), sensory feedback, and supraspinal drive signals from the brain.  Computational models of CPGs have been widely used for investigating the spinal cord contribution to animal locomotion control in computational neuroscience and in bio-inspired robotics. However, the contribution of supraspinal drive to anticipatory behavior, i.e. motor behavior that involves planning ahead of time (e.g. of footstep placements), is not yet properly understood. In particular, it is not clear whether the brain modulates CPG activity and/or directly modulates muscle activity (hence bypassing the CPG) for accurate foot placements. In this paper, we investigate the interaction of supraspinal drive and a CPG in an anticipatory locomotion scenario that involves stepping over gaps. By employing deep reinforcement learning (DRL), we train a neural network policy that replicates the supraspinal drive behavior. This policy can either modulate the CPG dynamics, or directly change actuation signals to bypass the CPG dynamics. Our results indicate that the direct supraspinal contribution to the actuation signal is a key component for a high gap crossing success rate. However, the CPG dynamics in the spinal cord are beneficial for gait smoothness and energy efficiency. Moreover, our investigation shows that sensing the front feet distances to the gap is the most important and sufficient sensory information for learning gap crossing. Our results support the biological hypothesis that cats and horses mainly control the front legs for obstacle avoidance, and that hind limbs follow an internal memory based on the front limbs' information. Our method enables the quadruped robot to cross gaps of up to $\mathbf{20} $ {cm} ($\mathbf{50 \%}$ of body-length) without any explicit dynamics modeling or Model Predictive Control (MPC). 
\end{abstract}

%\balance
%%%%%%%%%%%%%%%%%%%%%%%%%%%%%%%%%%%%%%%%%%%%%%%%%%%%%%%%%%%%%%%%%%%%%%%%%%%%%%%%
\section{Introduction and Related Work}
\label{sec:INTRODUCTION}

\begin{figure}[h]
\centering
\includegraphics[scale=0.43, trim ={0.0cm 0.0cm 0.0cm 0.0cm},clip]{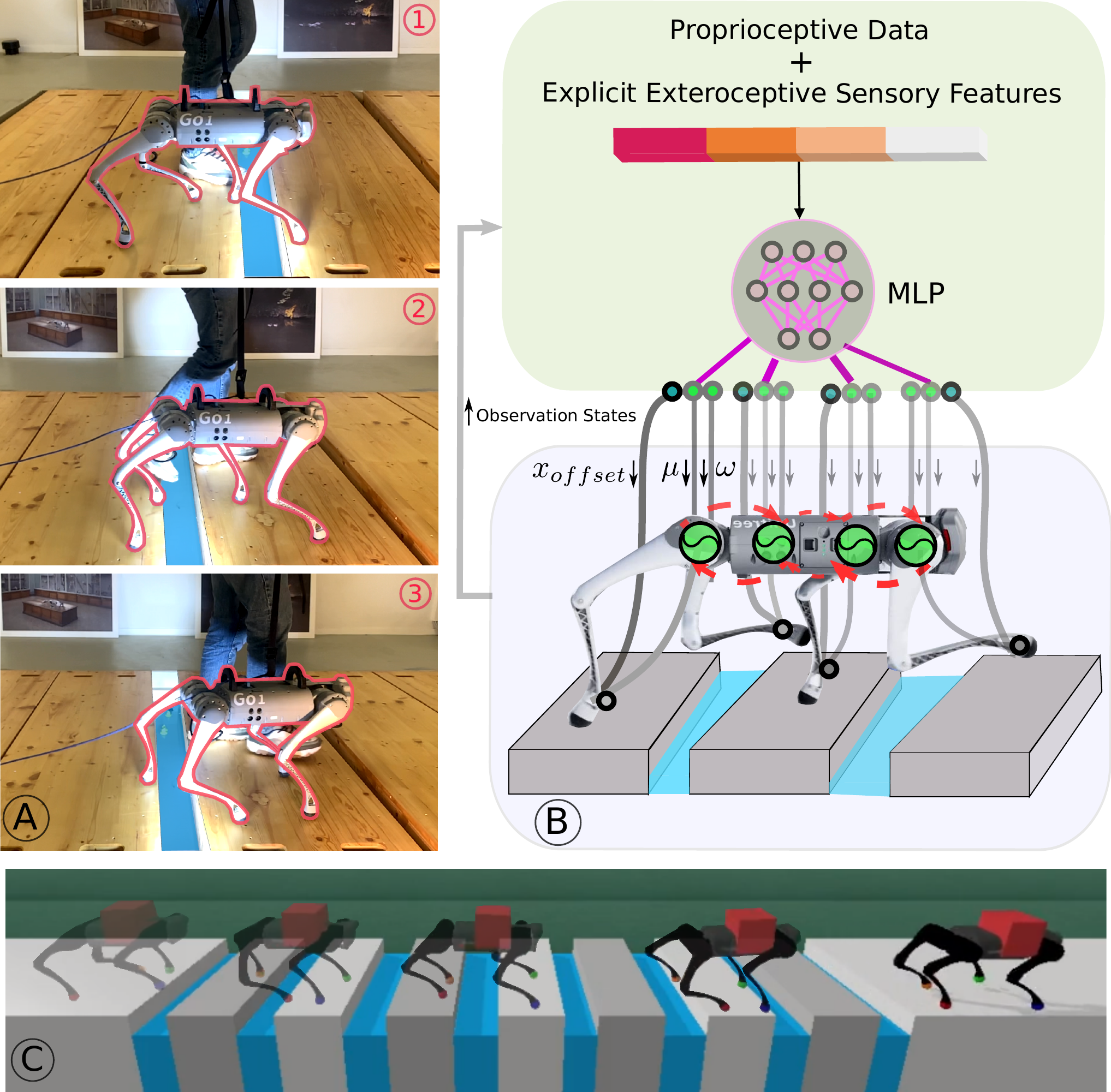}
      %trim={<left> <lower> <right> <upper>}
\vspace{-1.5em}
\caption{
\textbf{A}: Crossing a $15$ \si{cm} gap with Unitree Go1. \textbf{B}:  We represent the motor control system as a Puppeteer and Marionette,  where the supraspinal higher control centers work as a Puppeteer to manipulate the movement of the body (Marionette) with limited strings. The supraspinal drive controls movement by either modulating the frequency and amplitude of the CPG oscillators,
or directly sending actuation signals to bypass the CPG dynamics. \textbf{C}: Testing policy robustness by crossing variable gaps between 14 and 20 \si{cm} with an unknown 5 \si{kg} load. Videos: \insertWebPageLink }%\href{https://miladshafiee.github.io/puppeteer-and-marionette/}{{https://miladshafiee.github.io/puppeteer-and-marionette/}}  }

\vspace{-2.0em}
     \label{fig:puppet}
\end{figure}

Quadruped animals can perform highly agile motions including running, jumping over hurdles, and leaping over gaps.
Performing such anticipatory motor behaviors at high speeds requires complex interactions between the supraspinal drive, spinal cord dynamics, and sensory feedback~\cite{grillner2020current}. 
Through recent advances in machine learning and optimal control, animals' legged robot counterparts are becoming increasingly capable of traversing complex terrains~\cite{miki2022learning}. Robots can also be used as scientific tools to investigate biological hypotheses of animal adaptive behavior~\cite{ijspeert2014biorobotics}, and conversely we can take inspiration from the underlying mechanisms of animal locomotion to develop robotic systems that approach the agility of animals~\cite{sprowitz2013cheetah, hyun2014trot}. In this paper, by leveraging recent robotics tools, we investigate the interaction between the supraspinal drive, the central pattern generator (CPG), and sensory information to generate anticipatory locomotion control for a gap crossing task. We propose a hierarchical biologically-inspired framework, where higher control centers in the brain (represented by an artificial neural network) send supraspinal drive signals to either modulate the CPG dynamics, or directly output actuation signals which bypass the CPG dynamics.  

\subsection{Central Pattern Generators} 
It is widely accepted that the mammalian spinal cord contains a central pattern generator (CPG) that can produce basic locomotor rhythm in the absence of input from supraspinal drive and peripheral sensory feedback~\cite{grillner1978initiation}.
In robotics, abstract models of CPGs are commonly used for locomotion pattern generation~\cite{sprowitz2013cheetah,mos2013cat,aoi2017adaptive,kimura2007adaptive}, as well as to investigate biological hypotheses~\cite{ijspeert2007salamander, thandiackal2021emergence}. 
Besides the intrinsic oscillatory behavior of CPGs, several other properties such as robustness and implementation simplicity make CPGs desirable for locomotion control~\cite{ijspeert2008}. For legged robots, the CPG is usually designed for feedforward rhythm generation, and dynamic balancing is achieved with optimization~\cite{sprowitz2013cheetah} or hand-tuned feedback~\cite{righetti2008pattern, mos2013cat}. 
CPGs also provide an intuitive formulation for specifying different gaits~\cite{dutta2019programmable}, and spontaneous gait transitions can arise by increasing descending drive signals and incorporating contact force feedback~\cite{owaki2017quadruped,fukuoka2015simple} or vestibular feedback~\cite{fukui2019autonomous}.
In addition to proprioceptive feedback, incorporating exteroceptive feedback information allows for CPG-based locomotion over uneven terrains~\cite{saputra2021combining,gay2013learncpg} and navigation in complex environments~\cite{thor2022versatile,thor2020generic}.

Most of these studies consider the CPG to be isolated from higher control centers located in the brain. However, the interaction between supraspinal drive and the Central Pattern Generator during learning and planning leads to fascinating anticipatory locomotion in animals (i.e.~motor behavior that involves planning ahead of time). In this article, we investigate the roles of supraspinal drive and the CPG in learning such anticipatory behaviors.

\subsection{Learning Legged Locomotion}
Deep Reinforcement Learning (DRL) has emerged as a powerful approach for training robust legged locomotion control policies in simulation, and deploying these sim-to-real on hardware~\cite{iscen2018policies,hwangbo2019anymal,lee2020anymal,miki2022learning,siekmann2021blind,peng2020laikagoimitation,kumar2021rma,ji2022concurrent,margolis2022rapid,bellegarda2021robust,bellegarda2022cpgrl}. To facilitate this sim-to-real transfer, a variety of techniques can be employed such as online parameter adaptation~\cite{peng2020laikagoimitation,kumar2021rma}, learned state estimation modules~\cite{ji2022concurrent}, teacher-student training~\cite{kumar2021rma,lee2020anymal,miki2022learning}, and careful Markov Decision Process choices~\cite{bellegardaIROS19TaskSpaceRL,bellegarda2020robust,bellegarda2021robust,bellegarda2022cpgrl}. Most of these works view the trained artificial neural network (ANN) as a ``brain'' which has full access to complete whole-body proprioceptive sensing, which it queries at a high rate to update motor commands. Therefore, different gaits emerge through the combination of reward function tuning (i.e.~minimizing energy consumption~\cite{fu2022energy}), incorporating phase biases~\cite{shao2022gait,yang2022fast}, or imitating animal reference data to replicate bio-inspired movements~\cite{peng2018deepmimic, peng2020laikagoimitation}. However, here and in CPG-RL~\cite{bellegarda2022cpgrl}, we represent the ANN as a higher level control center which sends descending drive signals to modulate the central pattern generator in the spinal cord, and map this rhythm generation network to a pattern formation layer. Moreover, here we study the interplay between the ANN modulating the CPG directly, or bypassing the CPG to directly control lower level circuits. 

Beyond ``blind'' terrain locomotion, recent works incorporate exteroceptive sensing in the learning loop, for example for obstacle avoidance~\cite{yang2022learning} or walking over rough terrain by employing height maps~\cite{miki2022learning,pmlr-v164-rudin22a}. Gap crossing has been demonstrated by employing MPC and dynamic models for motion planning during learning~\cite{yu2022visual,xie2021glide,lee2022pi,margolis2022pixels}. For difficult simulation tasks, curriculum learning is helpful for surmounting increasingly challenging terrain~\cite{xie2020allsteps}, and jumping over large hurdles has been demonstrated by employing a mentor during the learning process~\cite{iscen2021learning}.

\subsection{Contribution}
Despite advances in understanding motor control of mammalian locomotion~\cite{grillner2020current}, little is known about the emergence of anticipatory locomotion skills through the interaction of supraspinal drive and CPGs. In this work, we investigate two broad neuroscience research questions:

\begin{itemize}
    \item What are the plausible contributions of supraspinal drive and CPG circuits in the spinal cord for producing anticipatory locomotion skills?
    \item  What are the necessary sensory feedback features for {learning} anticipatory locomotion skills? 
\end{itemize}

Although these research questions are broad, we view this work as a starting point for leveraging robotics techniques and biological inspiration to investigate the interaction of supraspinal drive (from the brain) with the CPG. We employ CPG models and a neural network (NN) policy trained with deep reinforcement learning to investigate this interaction for a gap-crossing task. For the first question, our results indicate that the direct supraspinal contribution to the actuation signal is a key component for a high success rate.  However, the CPG dynamics in the spinal cord are beneficial for gait smoothness and energy efficiency. 

Regarding the second question, our investigation shows that the front foot distance to the gap is the most important visually-extracted sensory information to successfully cross variable gaps. Our results show that front limb information is sufficient for learning gap-crossing, and that DRL can learn to create and encode an internal kinematic model by combining proprioceptive information with the internal CPG states to modulate the hind leg motion for gap crossing. This supports the biological hypothesis that cats and horses control their front legs for obstacle avoidance, and that hind legs follow an internal memory based on front foot information~\cite{mcvea2007contextual,whishaw2009hind}. Furthermore, in contrast to previous robotics works, to the best of our knowledge, this is the first learning-based framework with gap-crossing abilities which does not have a dynamical model, MPC, curriculum, or mentor in the loop. This illustrates the versatility of the proposed framework, which requires minimum expert knowledge (i.e.~no model of the system dynamics or more traditional optimal control). 

The rest of this paper is organized as follows. Section~\ref{sec:cpg_intro} describes the CPG topology. Section~\ref{sec:RL} details the DRL framework and design of the Markov Decision Process. Section~\ref{sec:results} presents results and analysis regarding the two mentioned research questions, and a brief conclusion is given in Section~\ref{sec:CONCLUSION}.

\section{Central Pattern Generators}
\label{sec:cpg_intro}
The locomotor system of vertebrates is organized such that the spinal CPGs are responsible for producing basic rhythmic patterns, while higher-level centers (i.e.~the motor cortex, cerebellum, and basal ganglia) are responsible for modulating the resulting patterns according to environmental conditions~\cite{grillner2020current}. Rybak et al.~\cite{rybak2006modelling} propose that biological CPGs have a two-level functional organization, with a half-center rhythm generator (RG) that determines movement frequency, and pattern formation (PF) circuits that determine the exact shapes of muscle activation signals. Similar organizations have also been used in robotics, for example in our previous work~\cite{bellegarda2022cpgrl}, and in~\cite{fukuhara2018spontaneous}. 
\subsection{Rhythm Generator (RG) Layer}
We employ amplitude-controlled phase oscillators to model the RG layer of the CPG circuits in the spinal cord. Such oscillators have been successfully used for locomotion control of legged robots~\cite{sprowitz2013cheetah,ijspeert2007salamander,bellegarda2022cpgrl} with the following dynamics:
\vspace{-1em}
\begin{align}
\ddot{r}_i &= \alpha\left(\frac{\alpha}{4} \left(\mu_i - r_i \right) - \dot{r}_i \right) \label{eq:salamander_r} \\
\dot{\theta}_i &= \omega_i +\sum_{j}^{} r_j w_{ij} \sin(\theta_j - \theta_i - \phi_{ij}) \label{eq:salamander_theta} 
\end{align}
where $r_i$ is the amplitude of the oscillator, $\theta_i$ is the phase of the oscillator, $\mu_i$ and $\omega_i$ are the intrinsic amplitude and frequency, $\alpha$ is a positive constant representing the convergence factor. Couplings between oscillators are defined by the weights $w_{ij}$ and phase biases $\phi_{ij}$. In this paper, we use the oscillators without neural coupling ($w_{ij}=0$), and gaits (i.e. phase relationships between limbs) are thus determined by the supraspinal control policy. As in~\cite{bellegarda2022cpgrl}, we will investigate the modulation of the intrinsic amplitude and frequency ($\mu_i$ and $\omega_i$) for each limb as control signals for the CPG.

\subsection{Pattern Formation (PF) Layer}
\begin{figure}[t!]
\centering
\includegraphics[scale=0.6923, trim ={0.0cm 0.0cm 0.0cm 0.},clip]{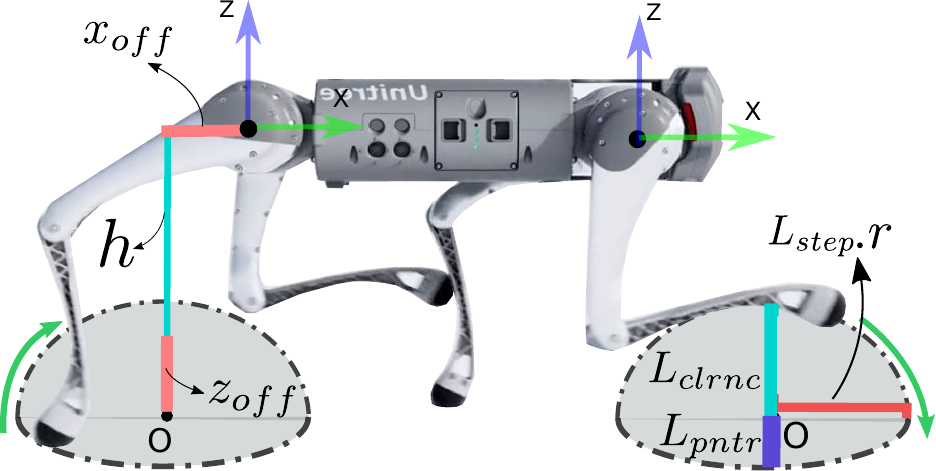}
      %trim={<left> <lower> <right> <upper>}
\vspace{-0.5em}
\caption{ Visualization of the task space foot trajectories generated by the PF layer.   
The oscillatory trajectory is built around a central point $O$. The offsets $x_{off}$ and $z_{off}$ are used to change the central point of oscillation. $x_{off}$ is a horizontal offset between the oscillation set-point and the center of the hip coordinate, and controlled directly by the supraspinal drive, bypassing the CPG dynamics.  $z_{off}+ h$ is the vertical distance between the oscillation set-point, $O$, and the center of the hip coordinate. $L_{step}r$ is the step length multiplied by the oscillator amplitude, $h$ is the nominal leg length, $L_{clrnc}$ is the max ground clearance during leg swing phase, and $L_{pntr}$ is the max ground penetration during stance. }
\vspace{-2em}
     \label{fig:feet-cpg}
\end{figure}

To map from the RG layer to joint commands, we first compute corresponding desired foot positions, and then calculate the desired joint positions with inverse kinematics. The desired foot position coordinates are formed as follows:
\vspace{-0.7em}
\begin{align}
x_{i,\text{foot}} &= \ \ x_{off, i}  -L_{step} (r_i) \cos(\theta_i) \label{eq:feet-task1} \\
z_{i,\text{foot}} &= \begin{cases}
    z_{off, i}-h+ L_{clrnc}\sin(\theta_i) & \text{if } \sin(\theta_i) > 0 \\
    z_{off, i}-h+L_{pntr}\sin(\theta_i) & \text{otherwise}
\end{cases} 
\label{eq:feet-task}
\end{align}

\noindent where $L_{step}$ is the step length, $h$ is the nominal leg length, $L_{clrnc}$ is the max ground clearance during swing, $L_{pntr}$ is the max ground penetration during stance, and $x_{off}$ and $z_{off}$ are set-points that change the equilibrium point of oscillation in the $x$ and $z$ directions. Modulating the foot horizontal offset $x_{off}$ and vertical offset $z_{off}$ represents direct supraspinal control of the general position of the limb, bypassing the rhythm generation layer. A description and visualization of the foot trajectory is illustrated in Figure~\ref{fig:feet-cpg}. 
\section{Hierarchical Bio-Inspired Learning of Anticipatory Gap Crossing Tasks} 
\label{sec:RL}

In this section we describe our hierarchical bio-inspired learning framework for learning anticipatory gap crossing abilities for quadruped robots. 
We represent the supraspinal controller as an artificial neural network 
which is trained with DRL to modulate both the feet positions (CPG offsets) and/or the intrinsic frequencies and amplitudes of oscillation for each limb to produce anticipatory behavior. 
The problem is represented as a Markov Decision Process (MDP), and we describe each of its components below.

\subsection{Action Space}
\label{sec:action}
We consider one RG layer for each limb based on  Equations~(\ref{eq:salamander_r}) and~(\ref{eq:salamander_theta}), where the RG output will be used in a PF layer to generate the spatio-temporal foot trajectories in Cartesian space (Equations~(\ref{eq:feet-task1}) and~(\ref{eq:feet-task})). We do not consider any explicit neural coupling (i.e.~$w_{ij}=0$), with the intuition that inter-limb coordination will be managed by the supraspinal drive. 

As in~\cite{bellegarda2022cpgrl}, our action space modulates the intrinsic amplitudes and frequencies of the CPG, by continuously updating $\mu_i$ and $\omega_i$ for each leg. However, unlike~\cite{bellegarda2022cpgrl}, we also consider modulating the oscillation set-points by directly learning foot Cartesian offsets $x_{off_i}, z_{off_i}$ for each leg. Thus, our action space can be summarized as~$\mathbf{a} = [\bm{\mu}, \bm{\omega}, \bm{x_{off}}, \bm{z_{off}} ] \in \mathbb{R}^{16}$. We divide the descending drive modulation into two categories: oscillatory components of the CPG dynamics $\mathbf{a}_{osc} = [\bm{\mu}, \bm{\omega}] \in \mathbb{R}^{8} $, and offset components $\mathbf{a}_{off} =  [\bm{x_{off}},\bm{z_{off}} ] \in \mathbb{R}^{8}$, shown in Equations (\ref{eq:salamander_r})-(\ref{eq:feet-task}). This separation allows us to investigate how gap crossing can best be accomplished, i.e.~by modulating CPG activity (by changing $\mu_i$ and $\omega_i$) and/or by directly updating the limb posture (by changing $x_{off_i}$ or $z_{off_i}$). Based on this investigation, we use $\mathbf{a} = [\bm{\mu}, \bm{\omega}, \bm{x_{off}}] \in \mathbb{R}^{12}$ for analyzing the roles of sensory feedback features in Section~\ref{subsec:sensory}. The agent selects these parameters at 100 Hz, which will therefore vary during each step according to sensory inputs. We use the following limits for each input during training: $\mu\in[0.5, 4]$, $\omega\in[0, 5]$ Hz, $x_{off_i}\in[-7, 7] cm$, $z_{off_i}\in[-7, 7] cm$.  

\begin{figure*}[h]
\centering
\includegraphics[scale=0.77, trim ={0.0cm 0.cm 0.0cm 0.0cm},clip]{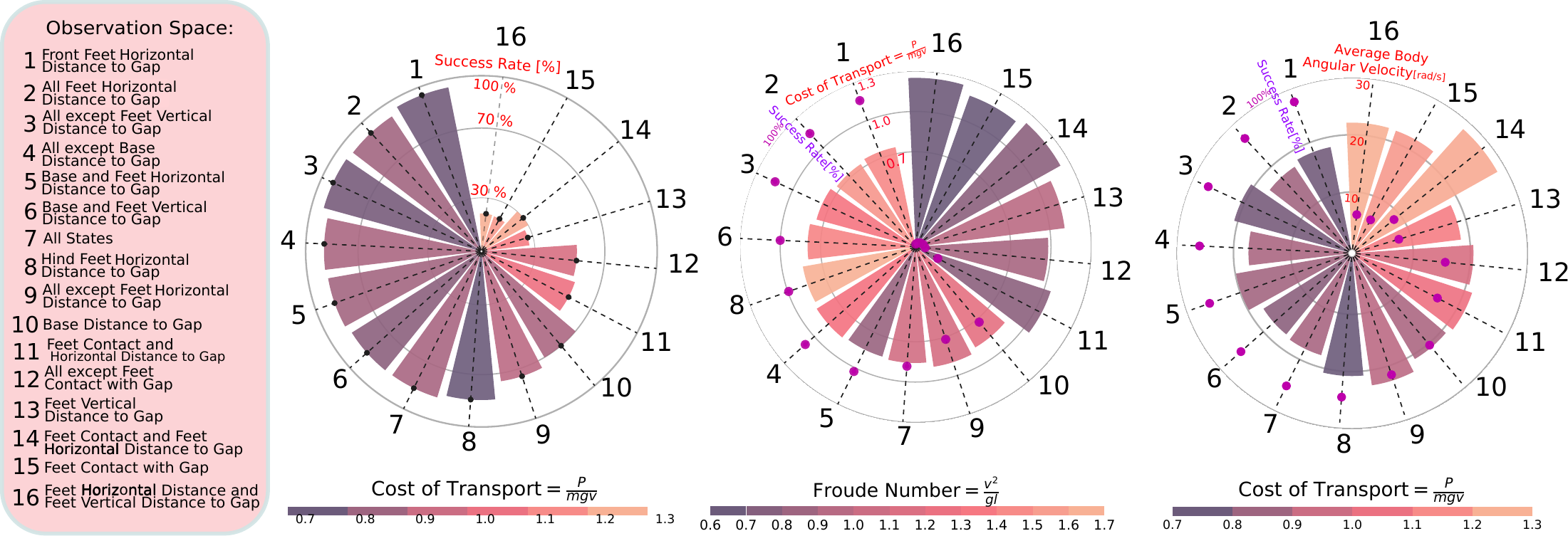}
      %trim={<left> <lower> <right> <upper>}
\vspace{-0.6em}
\caption{
Quantitative results from testing 16 policies trained with different combinations of  exteroceptive feedback features consisting of feet distance to the gap, base distance to the gap, feet vertical distance with the ground, and contact/penetration with the gap. All policies also include the flat terrain (proprioceptive sensing) observation, and the action space consists of both oscillatory and offset terms. We report the average results of testing the polices for 4000 samples each (100 attempts of crossing 7 gaps with randomized lengths between $[14,20]$ $cm$).  We characterize the viability performance of the system by the success rate, which is the proportion of the gaps in front of the robot that it could successfully cross.  Energy efficiency is characterized by the Cost of Transport (CoT), {mean velocity is characterized by the Froude number, and gait smoothness is evaluated by the mean angular velocity}.
}
\vspace{-1.6em}
     \label{fig:obs_comparison}
\end{figure*} 

\subsection{Observation Space}
\label{sec:obs_space}
We consider two different observation space types based on (1) only proprioceptive sensing (enough for locomotion on flat terrain) and (2) also including exteroceptive anticipatory features. Various exteroceptive anticipatory features will be investigated to understand the roles and importance of different sensory quantities.

\noindent \textbf{Flat terrain observation:}
We consider body  orientation, body linear and angular velocity, joint positions and velocities, foot contact booleans, the previous action chosen by the policy network, and CPG states $\{\bm{r,\dot{r},\theta,\dot{\theta}}\}$ as the flat terrain (proprioceptive sensing) observation space.
 
\noindent \textbf{{Exteroceptive Feedback Features}:}
We assume that the visual system and brain can extract important geometrical information such as foot distance to a gap, and we call such information \textit{exteroceptive feedback features}. We are interested in investigating which exteroceptive feedback features are most useful for the emergence of anticipatory locomotion skills. To reverse engineer this process, we divide the exteroceptive feedback features into two categories: \textit{predictive} and \textit{instantaneous} feedback  features. \textit{Predictive} features consist of foot distance and/or base distance to the beginning and end of a gap. \textit{Instantaneous} feedback features consist of boolean indicators of stepping into a gap (foot contact/penetration into the gap), as well as vertical distance of the foot with the ground (larger values over a gap). These features are instantaneous feedback, so they cannot be used to predict information about upcoming gaps. 

\subsection{Reward Function} 
\label{sec:reward}
Our reward function promotes forward progress, gap crossing ability, maintaining base stability, and energy efficiency with the following terms: 
\begin{equation}
\begin{split}
    r = \alpha_1\cdot{\min}({f}_x, {d}_{max}) + ({S}_{gap} +  {n}_{gap}) +\alpha_3\cdot|{y}_{base}| \\ + \alpha_4\cdot||\bm{o}_{base} - \bm{o}_{zero} || +  \alpha_5\cdot|\bm{\tau} \cdot (\dot{\bm{q}}_t-\dot{\bm{q}}_{t-1})| 
\end{split}
\nonumber
\end{equation}
\begin{itemize}
\item \textit{Forward progress}: 
In the first term, ${f}_x$ corresponds to forward progress in the world (along the $x$-direction). We limit this term to avoid exploiting
simulator dynamics and achieving unrealistic speeds, where ${d}_{max}$ is the maximum 
distance the robot will be rewarded for moving forward during each control cycle ($\alpha_1 = 2$). 
\item \textit{Gap reward}:  The agent receives a sparse reward (${S}_{gap}=3$) if it crosses a gap, 
and a negative reward of $ {n}_{gap}=-0.03$ for each control cycle during which a foot penetrates below ground level into a gap. 
\item \textit{Base $y$ direction penalty}: The third term penalizes lateral deviation of the body ($\alpha_3 = -0.05$). 
\item \textit{Base orientation penalty}: The fourth  term penalizes non-zero body orientation ($\alpha_4 = -0.02$). 
\item \textit{Power}: The fifth term penalizes power in order to find energy efficient gaits, where $\bm{\tau}$ and $\dot{\bm{q}}$  are joint torques and velocities ($\alpha_5 = -0.00008$). 
\end{itemize}

\subsection{Training details}
\label{sec:training_details}
We use PyBullet \cite{pybullet} as our physics engine for training and simulation purposes, and the Unitree {A1 and Go1 quadruped robots~\cite{unitreeA1}}.  To train the policies, we use Proximal Policy Optimization (PPO) \cite{ppo}, and Table \ref{table:RL} lists the PPO hyperparameters and neural network architecture.  The control frequency of the policy is 100 Hz, and the torques computed from the desired joint positions are updated at 1 kHz. The equations for each of the oscillators (Equations~\ref{eq:salamander_r} and~\ref{eq:salamander_theta}) are thus also integrated at 1 kHz. The joint PD controller gains are $K_p=100$ and $ K_d=2$.  All policies are trained for $3.5\times10^7$ samples.

\begin{table}[tpb]
\centering
\caption{PPO Hyperparameters and neural network architecture.}
\vspace{-0.7em}
\begin{tabular}{ c c | c c }
Parameter & Value &Parameter & Value\\
\hline
Batch size &  4096 &  SGD Iterations & 10 \\
SGD Mini-bach size & 128 & Discount factor & 0.99\\
Desired KL-divergence $kl^*$ & 0.01 & Learning rate $\alpha$ & 0.0001\\ 
GAE discount factor & 0.95  & Hidden Layers  & 2 \\
Clipping Threshold & 0.2 & Nodes  & [256,256] \\
Entropy coefficient & 0.01& Activation & tanh 
 \\
\hline
\vspace{-3.5em}
\end{tabular} \\
\label{table:RL}
\end{table}
\section{Results}
\label{sec:results}
In  this  section,  we  present  results of our proposed framework  for quadruped gap crossing scenarios. In Section~\ref{subsec:actuation}, we investigate the role of the supraspinal drive and  CPG  based on three criteria: success rate, energy efficiency, and gait smoothness. Furthermore, we investigate the effects of including varying exteroceptive sensory features on these criteria in Section~\ref{subsec:sensory}. Section~\ref{subsec:challenging} presents results for a more challenging task of crossing successive narrowly-spaced gaps, and Section~\ref{subsec:hw_result} discusses sim-to-real hardware results. The reader is encouraged to watch the  supplementary video for clear visualizations of all discussed experiments.

\subsection{ Contribution of CPG and Supraspinal Drive to Actuation }
 \label{subsec:actuation}
 In this section we train locomotion policies for the following scenarios and action spaces:
 \begin{itemize}
     \item[1.] Flat terrain. CPG only in $xz$ directions. 
     \item[2.] Gap terrain. CPG only in $xz$ directions. 
     \item[3.] Gap terrain. CPG in $z$ direction and offset in $x$ direction.
     \item[4.] Gap terrain. CPG in $xz$  and offset in $x$ direction.
     \item[5.] Gap terrain. Offsets only in $xz$ directions.
     \item[6.] Gap terrain. CPG and offsets both in $xz$ directions. 
 \end{itemize}
 The CPG in these cases means the agent (supraspinal drive) modulates the frequency and amplitude of Equations~(\ref{eq:salamander_r}) and~(\ref{eq:salamander_theta}).  The offset terms are considered as a part of the actuation signal applied directly by the supraspinal drive, bypassing the spinal cord dynamics. Cases 5 and 6 are the only cases in which we modulate the offset in the $z$ direction. 
 
We train policies for each case for $3.5 \times 10^7$ samples for episodes of $10$ $s$ on terrains with 7 consecutive gaps (except for Case 1, which is trained on flat terrain only) with all exteroceptive feedback features in the observation space. Each gap length is randomized in $[14,20]$ $cm$ during both training and test time, with $1.4$ $m$ distances between gaps. An episode terminates early because of a fall,~i.e. if the body height drops below $15$ $cm$. We define the success rate as the number of gaps successfully crossed out of the total number of gaps. In order to test the six policies, we perform 30 policy rollouts on a test environment of locomoting over 7 randomized gaps. Table~\ref{table:action-space} summarizes the results from investigating how supraspinal drive can modulate locomotion in these six cases. 

\subsubsection{{Gap Crossing Success Rate}}
Case 5 has the highest success rate of $99 \%$, indicating the benefit of direct supraspinal actuation in anticipatory scenarios. Case 4, with both oscillatory and offset terms in the $x$ direction, has the second highest success rate of $97 \%$. The third highest success rate is for Case 6 with both oscillatory and offset terms in both $x$ and $z$ directions.  The fourth highest success rate is Case 3 (with only the offset component in the $x$ direction), and Case 2 (with only the CPG components) has the fifth best success rate of $17\%$. These results show that  direct  supraspinal actuation of the foot offset/position  is critical for successful gap-crossing, though the CPG can contribute to a high success rate in the absence of $z$ offset modulation.
\subsubsection{{Gait Smoothness}}
To compare the gait smoothness between policies, we analyze the robot body oscillations during locomotion, and in particular the average angular velocity of the robot body $\bar\omega_{Body}=({\sum_{t=1}^{N} \abs{\omega_{x,t}}+\abs{\omega_{y,t}}+\abs{\omega_{z,t}}})/ (3 N)$.

Body orientation deviations are penalized in the reward function, as high (absolute) angular velocities tend to correspond to shaky gait patterns. 
As shown in Table \ref{table:action-space}, the first case has the smoothest gait. This is expected since it corresponds to steady-state locomotion behavior on flat terrain. A comparison of the third and fourth cases indicates a $45 \%$ reduction in body oscillation {when the agent can also modulate CPG amplitudes}. The gait smoothness of case 5 is drastically reduced by removing the CPG dynamics. This result shows the importance of spinal cord dynamics (limit-cycle oscillatory dynamics) for obtaining smooth locomotion.

\subsubsection{{Cost of Transport (CoT) and Froude number}}
We investigate gait efficiency by comparing the CoT, and mean velocity with the Froude number~\cite{sprowitz2013cheetah}.  We observe that the fourth case (with both oscillatory and offset components), has the best combined CoT, Froude number, and gait smoothness (low $\bar\omega_{Body}$).  This demonstrates the benefit of having both supraspinal drive and CPG dynamics for coordinating locomotion. Case 3,  which has the CPG oscillatory component in $z$ and offset in $x$, has the lowest CoT, and we observe significant added energy expenditure by removing the CPG dynamics (Case 5). Case 6 shows that having both oscillatory and offset terms leads to the highest Froude number, but also a high CoT and $\bar\omega_{Body}$, suggesting that overparameterizing the action space can make it difficult for the agent to converge to an optimal policy. 

\begin{figure*}[h]
\centering
\includegraphics[scale=0.783, trim ={0.0cm 0.0cm 0.0cm 0.0cm},clip]{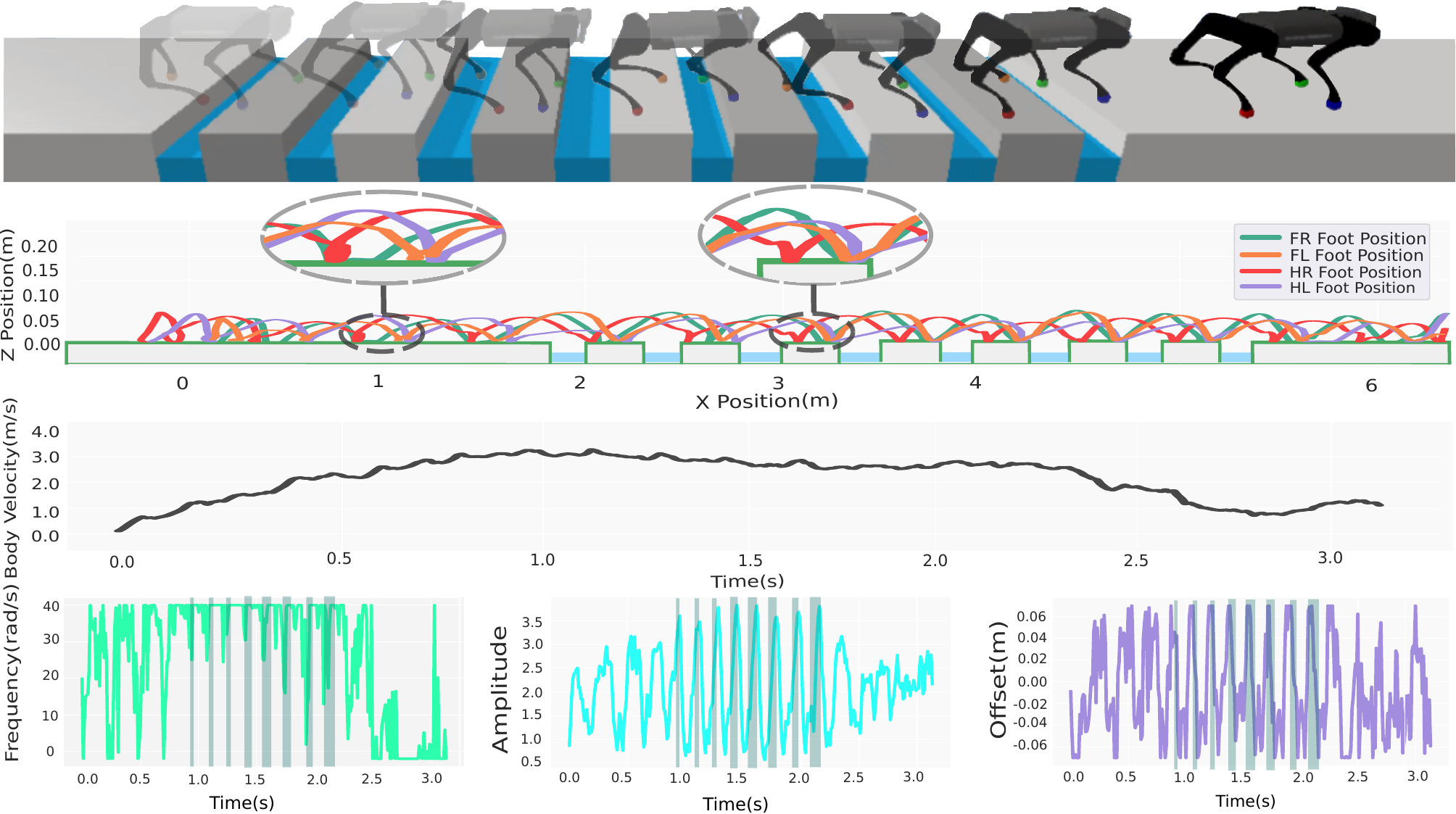}
      %trim={<left> <lower> <right> <upper>}
\vspace{-0.4em}
\caption{ Crossing 8 gaps with randomized lengths between $[14,20]$ cm, with only 30 cm contact surfaces. \textbf{Top:} simulation snapshots. \textbf{Middle:} body velocity and foot positions in the XZ plane.  \textbf{Bottom:} CPG frequency, amplitude, and offset for the front left limb. The shadow bars indicate when the foot is over a gap. }
\vspace{-2.0em}
     \label{fig:Gapsnapshot}
\end{figure*}

 \subsection{Roles of Feedback Features for Anticipatory Locomotion } 
 \label{subsec:sensory}
In this section we investigate which exteroceptive sensory feedback  information is necessary and sufficient for learning and planning anticipatory tasks. As shown in Figure~\ref{fig:obs_comparison}, we consider 16 different combinations of predictive and instantaneous feedback features as described in Section \ref{sec:obs_space}. We train 16 different policies with the Case 4 action space from Section~\ref{subsec:actuation} (i.e.~with CPG and $x$ offset modulation). 

\begin{table}[tpb]
\centering
\vspace{0.06in}
\caption{Testing policies trained with different combinations of oscillatory and offset terms in the action space. $\bar\omega_{Body}$ is the average body angular velocity.  Case 1 is for walking on flat terrain without gaps. We only show mean values since the standard deviations are small (i.e.~less than 10$\%$ of the means).}
\vspace{-0.5em}
\resizebox{0.99\linewidth}{!}{
\begin{tabular}{ c  c  c c c c  } 
Case &  $x_{osc}$ $x_{off}$ $z_{osc}$  $z_{off}$   &Success[$\%$]   &  CoT & Froude   & $\bar\omega_{Body}$ 
  \\ 
\hline
1  & $\checkmark$ \hspace{7pt}  $\times$ \hspace{7pt} $\checkmark$ \hspace{4pt} $\times$& $\times$   & 0.92 & 0.34  & \textbf{0.36}\\
2 & $\checkmark$  \hspace{7pt} $\times$  \hspace{7pt} $\checkmark$  \hspace{4pt}  $\times$   & 17  & 1.45 & 0.29  & 0.73
\\
3 & $\times$ \hspace{7pt} $\checkmark$ \hspace{7pt} $\checkmark$ \hspace{4pt} $\times$  & 60 & \textbf{0.84} & 0.29 & 0.77 \\
4 & $\checkmark$ \hspace{7pt}  $\checkmark$ \hspace{7pt} $\checkmark$ \hspace{4pt} $\times$ &  97 & 0.94 & 0.55  & 0.42\\
5 & $\times$ \hspace{7pt} $\checkmark$ \hspace{7pt} $\times$ \hspace{4pt} $\checkmark$ & \textbf{99} & 1.24 & 0.56  & 0.96\\
6 & $\checkmark$ \hspace{7pt} $\checkmark$ \hspace{7pt} $\checkmark$ \hspace{4pt} $\checkmark$ & 93 & 1.35 & \textbf{0.88}  & 0.85\\
\hline
\end{tabular}
}
\vspace{-2em}
\label{table:action-space}
\end{table}

For evaluation, we rollout each policy 100 times and present mean results across all tests. Fig.~\ref{fig:obs_comparison}-A shows the gap crossing success rate (by bar height) and CoT (by color). Our results show that policy 1, which has the front feet distances to the gap in the observation space, has both one of the best success rates as well as one of the lowest CoTs. These results show that front leg information is sufficient for learning the gap-crossing task.  As the agent is explicitly blind about hind leg positions, this forces it to learn an internal kinematic model by combining exteroceptive front feet positions to the gap, internal CPG states, and proprioceptive sensing to modulate the hind leg motions for gap crossing.  This result supports the biological hypothesis that cats and horses control the front legs for obstacle avoidance, and that hind legs follow based on an internal kinematic memory~\cite{mcvea2007contextual,whishaw2009hind}.

Notably, as could be expected, the 8 policies with the highest success rates contain the feet and/or base distances to the gap in the observation space, which indicates the importance of predictive feedback features in the observation space.  Interestingly, policy 7, which observes all discussed exteroceptive sensing in the observation space, has a slightly lower success rate  with respect to the first 6 policies with subsets of the full sensing. This suggests that including supererogatory information in the observation space may not necessarily improve the quality of the learned policy, and may even prevent the convergence of the RL algorithm in finding the optimal policy.

Figure~\ref{fig:obs_comparison}-B shows the CoT as the bar height, with the color indicating the Froude number. The lowest CoT and highest Froude number are for the first three policies, which include the front feet positions in the observation space. The first 10 policies show that having predictive information in the observation space helps to learn an energy-efficient gait for the gap-crossing task. 

Figure~\ref{fig:obs_comparison}-C shows the average body angular velocity to investigate gait smoothness. We observe that having feet distances to the gap in the observation space leads to lower average body angular velocities, and as a result smoother gaits. 

\subsection{Training for a More Challenging Gap-Crossing Scenario} 
\label{subsec:challenging}
In this section, we train the robot to cross 8 gaps with the front feet distances to the gap in the observation space (the same as policy 1 from Section~\ref{subsec:sensory}), with 30 $cm$ platforms between each gap. The gaps have randomized lengths between $[14,20]$ $cm$, and the first gap position is randomized between $[1.25,2.25]$ $m$. As shown in Figure \ref{fig:Gapsnapshot}, the supraspinal drive increases the velocity of the robot by increasing the frequency of the CPG. The desired velocity for the robot is 1 $m/s$, however, the agent has learned to increase the velocity of the robot to up to 3 $m/s$ to overcome the gaps. We observe that the policy increases the limb frequency to near its maximum limits for all legs  as soon as it reaches the first gap. This indicates that the supraspinal drive modulates the locomotion speed by increasing the CPG frequencies. The oscillation amplitude also changes the step position and reaches maximum values for each foot to cross the gaps.

As seen in bottom right of Figure \ref{fig:Gapsnapshot}, the offset term is an important component for inter-limb coordination and can explicitly modulate the step position. On average it has the highest and lowest values when the foot starts and stops crossing a gap. Figure \ref{fig:Gapsnapshot} (middle) interestingly shows the robot places its HL limb approximately where the FL limb was located in the previous stride. 

\subsection{Hardware Experiment} 
\label{subsec:hw_result}
We perform a sim-to-real transfer of policy 1 from Section~\ref{subsec:sensory} to the Go1 hardware for a two gap scenario with widths of $15$ $cm$ and $7$ $cm$. We simplify the sim-to-real transfer by using a trained neural network to capture the actuator dynamics~\cite{hwangbo2019anymal,margolis2022walk}, and we assume knowledge of the relative gap distance to the robot from an equivalent scenario completed in simulation. Figure~\ref{fig:puppet}-A shows snapshots of trotting over the gaps with a mean velocity of 0.7 $m/s$. 

\section{Conclusion}
\label{sec:CONCLUSION}
In this work, we have proposed a framework to investigate the interactions between supraspinal drive and the CPG to generate anticipatory quadruped locomotion in gap crossing scenarios. Our results show that supraspinal drive is critical for high success rates for gap crossing, but CPG dynamics are beneficial for energy efficiency and gait smoothness.  Moreover, our results show that the front foot distance to the gap is the most important and sufficient visually-extracted sensory information for learning gap crossing scenarios.  This supports the biological hypothesis that cats and horses control their front legs for obstacle avoidance, and that hind legs follow an internal memory based on the front feet information~\cite{mcvea2007contextual,whishaw2009hind}. This shows that DRL is able to create and encode an internal kinematic model with proprioceptive sensing to modulate the hind leg motion for gap crossing.  Furthermore, in contrast to previous work, to the best of our knowledge, this is the first RL framework with gap-crossing capability without having a dynamical model, MPC, curriculum, or mentor in the loop.  

\vspace{-0.5em}

\section*{Acknowledgements}
We would like to thank Alessandro Crespi for assisting with hardware setup. 

\bibliographystyle{IEEEtran}
\balance
\bibliography{root}

\begin{thebibliography}{10}
\providecommand{\url}[1]{#1}
\csname url@rmstyle\endcsname
\providecommand{\newblock}{\relax}
\providecommand{\bibinfo}[2]{#2}
\providecommand\BIBentrySTDinterwordspacing{\spaceskip=0pt\relax}
\providecommand\BIBentryALTinterwordstretchfactor{4}
\providecommand\BIBentryALTinterwordspacing{\spaceskip=\fontdimen2\font plus
\BIBentryALTinterwordstretchfactor\fontdimen3\font minus
  \fontdimen4\font\relax}
\providecommand\BIBforeignlanguage[2]{{%
\expandafter\ifx\csname l@#1\endcsname\relax
\typeout{** WARNING: IEEEtran.bst: No hyphenation pattern has been}%
\typeout{** loaded for the language `#1'. Using the pattern for}%
\typeout{** the default language instead.}%
\else
\language=\csname l@#1\endcsname
\fi
#2}}

\bibitem{grillner2020current}
S.~Grillner and A.~El~Manira, ``Current principles of motor control, with
  special reference to vertebrate locomotion,'' \emph{Physiological reviews},
  vol. 100, no.~1, pp. 271--320, 2020.

\bibitem{miki2022learning}
T.~Miki, J.~Lee, J.~Hwanbo, L.~Wellhausen, V.~Koltun, and M.~Hutter, ``Learning
  robust perceptive locomotion for quadrupedal robots in the wild,''
  \emph{Science Robotics}, 2022.

\bibitem{ijspeert2014biorobotics}
A.~J. Ijspeert, ``Biorobotics: Using robots to emulate and investigate agile
  locomotion,'' \emph{science}, vol. 346, no. 6206, pp. 196--203, 2014.

\bibitem{sprowitz2013cheetah}
A.~Spröwitz, A.~Tuleu, M.~Vespignani, M.~Ajallooeian, E.~Badri, and A.~J.
  Ijspeert, ``Towards dynamic trot gait locomotion: Design, control, and
  experiments with cheetah-cub, a compliant quadruped robot,'' \emph{The
  International Journal of Robotics Research}, vol.~32, no.~8, pp. 932--950,
  2013.

\bibitem{hyun2014trot}
D.~J. Hyun, S.~Seok, J.~Lee, and S.~Kim, ``High speed trot-running:
  Implementation of a hierarchical controller using proprioceptive impedance
  control on the mit cheetah,'' \emph{The International Journal of Robotics
  Research}, vol.~33, no.~11, pp. 1417--1445, 2014.

\bibitem{grillner1978initiation}
S.~Grillner and S.~Rossignol, ``On the initiation of the swing phase of
  locomotion in chronic spinal cats,'' \emph{Brain research}, vol. 146, no.~2,
  pp. 269--277, 1978.

\bibitem{mos2013cat}
M.~Ajallooeian, S.~Pouya, A.~Sproewitz, and A.~J. Ijspeert, ``Central pattern
  generators augmented with virtual model control for quadruped rough terrain
  locomotion,'' in \emph{2013 IEEE International Conference on Robotics and
  Automation}, 2013, pp. 3321--3328.

\bibitem{aoi2017adaptive}
S.~Aoi, P.~Manoonpong, Y.~Ambe, F.~Matsuno, and F.~W{\"o}rg{\"o}tter,
  ``Adaptive control strategies for interlimb coordination in legged robots: a
  review,'' \emph{Frontiers in neurorobotics}, vol.~11, p.~39, 2017.

\bibitem{kimura2007adaptive}
H.~Kimura, Y.~Fukuoka, and A.~H. Cohen, ``Adaptive dynamic walking of a
  quadruped robot on natural ground based on biological concepts,'' \emph{The
  International Journal of Robotics Research}, vol.~26, no.~5, pp. 475--490,
  2007.

\bibitem{ijspeert2007salamander}
A.~J. Ijspeert, A.~Crespi, D.~Ryczko, and J.-M. Cabelguen, ``From swimming to
  walking with a salamander robot driven by a spinal cord model,''
  \emph{Science}, vol. 315, no. 5817, pp. 1416--1420, 2007.

\bibitem{thandiackal2021emergence}
R.~Thandiackal, K.~Melo, L.~Paez, J.~Herault, T.~Kano, K.~Akiyama, F.~Boyer,
  D.~Ryczko, A.~Ishiguro, and A.~J. Ijspeert, ``Emergence of robust
  self-organized undulatory swimming based on local hydrodynamic force
  sensing,'' \emph{Science Robotics}, vol.~6, no.~57, 2021.

\bibitem{ijspeert2008}
A.~J. Ijspeert, ``Central pattern generators for locomotion control in animals
  and robots: A review,'' \emph{Neural Networks}, vol.~21, no.~4, pp. 642--653,
  2008, robotics and Neuroscience.

\bibitem{righetti2008pattern}
L.~Righetti and A.~J. Ijspeert, ``Pattern generators with sensory feedback for
  the control of quadruped locomotion,'' in \emph{2008 IEEE International
  Conference on Robotics and Automation}.\hskip 1em plus 0.5em minus
  0.4em\relax IEEE, 2008, pp. 819--824.

\bibitem{dutta2019programmable}
S.~Dutta, A.~Parihar, A.~Khanna, J.~Gomez, W.~Chakraborty, M.~Jerry,
  B.~Grisafe, A.~Raychowdhury, and S.~Datta, ``Programmable coupled oscillators
  for synchronized locomotion,'' \emph{Nature communications}, vol.~10, no.~1,
  pp. 1--10, 2019.

\bibitem{owaki2017quadruped}
D.~Owaki and A.~Ishiguro, ``A quadruped robot exhibiting spontaneous gait
  transitions from walking to trotting to galloping,'' \emph{Scientific
  reports}, vol.~7, no.~1, pp. 1--10, 2017.

\bibitem{fukuoka2015simple}
Y.~Fukuoka, Y.~Habu, and T.~Fukui, ``A simple rule for quadrupedal gait
  generation determined by leg loading feedback: a modeling study,''
  \emph{Scientific reports}, vol.~5, no.~1, pp. 1--11, 2015.

\bibitem{fukui2019autonomous}
T.~Fukui, H.~Fujisawa, K.~Otaka, and Y.~Fukuoka, ``Autonomous gait transition
  and galloping over unperceived obstacles of a quadruped robot with cpg
  modulated by vestibular feedback,'' \emph{Robotics and Autonomous Systems},
  vol. 111, pp. 1--19, 2019.

\bibitem{saputra2021combining}
A.~A. Saputra, J.~Botzheim, A.~J. Ijspeert, and N.~Kubota, ``Combining reflexes
  and external sensory information in a neuromusculoskeletal model to control a
  quadruped robot,'' \emph{IEEE Transactions on Cybernetics}, 2021.

\bibitem{gay2013learncpg}
S.~Gay, J.~Santos-Victor, and A.~Ijspeert, ``Learning robot gait stability
  using neural networks as sensory feedback function for central pattern
  generators,'' in \emph{2013 IEEE/RSJ International Conference on Intelligent
  Robots and Systems}, 2013, pp. 194--201.

\bibitem{thor2022versatile}
M.~Thor and P.~Manoonpong, ``Versatile modular neural locomotion control with
  fast learning,'' \emph{Nature Machine Intelligence}, vol.~4, no.~2, pp.
  169--179, 2022.

\bibitem{thor2020generic}
M.~Thor, T.~Kulvicius, and P.~Manoonpong, ``Generic neural locomotion control
  framework for legged robots,'' \emph{IEEE transactions on neural networks and
  learning systems}, vol.~32, no.~9, pp. 4013--4025, 2020.

\bibitem{iscen2018policies}
A.~Iscen, K.~Caluwaerts, J.~Tan, T.~Zhang, E.~Coumans, V.~Sindhwani, and
  V.~Vanhoucke, ``Policies modulating trajectory generators,'' in
  \emph{Conference on Robot Learning}.\hskip 1em plus 0.5em minus 0.4em\relax
  PMLR, 2018, pp. 916--926.

\bibitem{hwangbo2019anymal}
J.~Hwangbo, J.~Lee, A.~Dosovitskiy, D.~Bellicoso, V.~Tsounis, V.~Koltun, and
  M.~Hutter, ``Learning agile and dynamic motor skills for legged robots,''
  \emph{Science Robotics}, vol.~4, no.~26, 2019.

\bibitem{lee2020anymal}
J.~Lee, J.~Hwangbo, L.~Wellhausen, V.~Koltun, and M.~Hutter, ``Learning
  quadrupedal locomotion over challenging terrain,'' \emph{Science Robotics},
  vol.~5, no.~47, 2020.

\bibitem{siekmann2021blind}
J.~Siekmann, K.~Green, J.~Warila, A.~Fern, and J.~Hurst, ``Blind bipedal stair
  traversal via sim-to-real reinforcement learning,'' \emph{arXiv preprint
  arXiv:2105.08328}, 2021.

\bibitem{peng2020laikagoimitation}
X.~B. Peng, E.~Coumans, T.~Zhang, T.-W. Lee, J.~Tan, and S.~Levine, ``Learning
  agile robotic locomotion skills by imitating animals,'' 2020.

\bibitem{kumar2021rma}
A.~Kumar, Z.~Fu, D.~Pathak, and J.~Malik, ``Rma: Rapid motor adaptation for
  legged robots,'' \emph{arXiv preprint arXiv:2107.04034}, 2021.

\bibitem{ji2022concurrent}
G.~Ji, J.~Mun, H.~Kim, and J.~Hwangbo, ``Concurrent training of a control
  policy and a state estimator for dynamic and robust legged locomotion,''
  \emph{IEEE Robotics and Automation Letters}, vol.~7, no.~2, pp. 4630--4637,
  2022.

\bibitem{margolis2022rapid}
G.~B. Margolis, G.~Yang, K.~Paigwar, T.~Chen, and P.~Agrawal, ``Rapid
  locomotion via reinforcement learning,'' \emph{arXiv preprint
  arXiv:2205.02824}, 2022.

\bibitem{bellegarda2021robust}
G.~Bellegarda, Y.~Chen, Z.~Liu, and Q.~Nguyen, ``Robust high-speed running for
  quadruped robots via deep reinforcement learning,'' in \emph{2022 IEEE/RSJ
  International Conference on Intelligent Robots and Systems (IROS)}, 2022, pp.
  10\,364--10\,370.

\bibitem{bellegarda2022cpgrl}
G.~Bellegarda and A.~Ijspeert, ``{CPG-RL}: Learning central pattern generators
  for quadruped locomotion,'' \emph{IEEE Robotics and Automation Letters},
  vol.~7, no.~4, pp. 12\,547--12\,554, 2022.

\bibitem{bellegardaIROS19TaskSpaceRL}
G.~{Bellegarda} and K.~{Byl}, ``Training in task space to speed up and guide
  reinforcement learning,'' in \emph{2019 IEEE/RSJ International Conference on
  Intelligent Robots and Systems (IROS)}, 2019, pp. 2693--2699.

\bibitem{bellegarda2020robust}
G.~Bellegarda and Q.~Nguyen, ``Robust quadruped jumping via deep reinforcement
  learning,'' \emph{arXiv preprint arXiv:2011.07089}, 2020.

\bibitem{fu2022energy}
Z.~Fu, A.~Kumar, J.~Malik, and D.~Pathak, ``Minimizing energy consumption leads
  to the emergence of gaits in legged robots,'' in \emph{Proceedings of the 5th
  Conference on Robot Learning}, ser. Proceedings of Machine Learning Research,
  A.~Faust, D.~Hsu, and G.~Neumann, Eds., vol. 164.\hskip 1em plus 0.5em minus
  0.4em\relax PMLR, 08--11 Nov 2022, pp. 928--937.

\bibitem{shao2022gait}
Y.~Shao, Y.~Jin, X.~Liu, W.~He, H.~Wang, and W.~Yang, ``Learning free gait
  transition for quadruped robots via phase-guided controller,'' \emph{IEEE
  Robotics and Automation Letters}, vol.~7, no.~2, pp. 1230--1237, 2022.

\bibitem{yang2022fast}
Y.~Yang, T.~Zhang, E.~Coumans, J.~Tan, and B.~Boots, ``Fast and efficient
  locomotion via learned gait transitions,'' in \emph{Conference on Robot
  Learning}.\hskip 1em plus 0.5em minus 0.4em\relax PMLR, 2022, pp. 773--783.

\bibitem{peng2018deepmimic}
X.~B. Peng, P.~Abbeel, S.~Levine, and M.~van~de Panne, ``Deepmimic:
  Example-guided deep reinforcement learning of physics-based character
  skills,'' \emph{ACM Transactions on Graphics (TOG)}, vol.~37, no.~4, pp.
  1--14, 2018.

\bibitem{yang2022learning}
R.~Yang, M.~Zhang, N.~Hansen, H.~Xu, and X.~Wang, ``Learning vision-guided
  quadrupedal locomotion end-to-end with cross-modal transformers,'' in
  \emph{International Conference on Learning Representations}, 2022.

\bibitem{pmlr-v164-rudin22a}
N.~Rudin, D.~Hoeller, P.~Reist, and M.~Hutter, ``Learning to walk in minutes
  using massively parallel deep reinforcement learning,'' in \emph{Proceedings
  of the 5th Conference on Robot Learning}, ser. Proceedings of Machine
  Learning Research, A.~Faust, D.~Hsu, and G.~Neumann, Eds., vol. 164.\hskip
  1em plus 0.5em minus 0.4em\relax PMLR, 08--11 Nov 2022, pp. 91--100.

\bibitem{yu2022visual}
W.~Yu, D.~Jain, A.~Escontrela, A.~Iscen, P.~Xu, E.~Coumans, S.~Ha, J.~Tan, and
  T.~Zhang, ``Visual-locomotion: Learning to walk on complex terrains with
  vision,'' in \emph{Conference on Robot Learning}.\hskip 1em plus 0.5em minus
  0.4em\relax PMLR, 2022, pp. 1291--1302.

\bibitem{xie2021glide}
Z.~Xie, X.~Da, B.~Babich, A.~Garg, and M.~van~de Panne, ``Glide: Generalizable
  quadrupedal locomotion in diverse environments with a centroidal model,''
  \emph{arXiv preprint arXiv:2104.09771}, 2021.

\bibitem{lee2022pi}
K.-H. Lee, O.~Nachum, T.~Zhang, S.~Guadarrama, J.~Tan, and W.~Yu, ``Pi-ars:
  Accelerating evolution-learned visual-locomotion with predictive information
  representations,'' \emph{arXiv preprint arXiv:2207.13224}, 2022.

\bibitem{margolis2022pixels}
G.~B. Margolis, T.~Chen, K.~Paigwar, X.~Fu, D.~Kim, S.~b. Kim, and P.~Agrawal,
  ``Learning to jump from pixels,'' in \emph{Proceedings of the 5th Conference
  on Robot Learning}, ser. Proceedings of Machine Learning Research, vol.
  164.\hskip 1em plus 0.5em minus 0.4em\relax PMLR, 08--11 Nov 2022, pp.
  1025--1034.

\bibitem{xie2020allsteps}
Z.~Xie, H.~Y. Ling, N.~H. Kim, and M.~van~de Panne, ``Allsteps:
  Curriculum-driven learning of stepping stone skills,'' in \emph{Computer
  Graphics Forum}, vol.~39, no.~8.\hskip 1em plus 0.5em minus 0.4em\relax Wiley
  Online Library, 2020, pp. 213--224.

\bibitem{iscen2021learning}
A.~Iscen, G.~Yu, A.~Escontrela, D.~Jain, J.~Tan, and K.~Caluwaerts, ``Learning
  agile locomotion skills with a mentor,'' in \emph{2021 IEEE International
  Conference on Robotics and Automation (ICRA)}.\hskip 1em plus 0.5em minus
  0.4em\relax IEEE, 2021, pp. 2019--2025.

\bibitem{mcvea2007contextual}
D.~McVea and K.~Pearson, ``Contextual learning and obstacle memory in the
  walking cat,'' \emph{Integrative and Comparative Biology}, vol.~47, no.~4,
  pp. 457--464, 2007.

\bibitem{whishaw2009hind}
I.~Q. Whishaw, L.-A.~R. Sacrey, and B.~Gorny, ``Hind limb stepping over
  obstacles in the horse guided by place-object memory,'' \emph{Behavioural
  brain research}, vol. 198, no.~2, pp. 372--379, 2009.

\bibitem{rybak2006modelling}
I.~A. Rybak, N.~A. Shevtsova, M.~Lafreniere-Roula, and D.~A. McCrea,
  ``Modelling spinal circuitry involved in locomotor pattern generation:
  insights from deletions during fictive locomotion,'' \emph{The Journal of
  physiology}, vol. 577, no.~2, pp. 617--639, 2006.

\bibitem{fukuhara2018spontaneous}
A.~Fukuhara, D.~Owaki, T.~Kano, R.~Kobayashi, and A.~Ishiguro, ``Spontaneous
  gait transition to high-speed galloping by reconciliation between body
  support and propulsion,'' \emph{Advanced robotics}, vol.~32, no.~15, pp.
  794--808, 2018.

\bibitem{pybullet}
E.~Coumans and Y.~Bai, ``Pybullet, a python module for physics simulation for
  games, robotics and machine learning,'' \url{http://pybullet.org},
  2016--2019.

\bibitem{unitreeA1}
\BIBentryALTinterwordspacing
{Unitree Robotics}. (2021, February) A1. [Online]. Available:
  \url{https://www.unitree.com/products/a1/}
\BIBentrySTDinterwordspacing

\bibitem{ppo}
J.~Schulman, F.~Wolski, P.~Dhariwal, A.~Radford, and O.~Klimov, ``Proximal
  policy optimization algorithms,'' \emph{CoRR}, vol. abs/1707.06347, 2017.

\bibitem{margolis2022walk}
G.~B. Margolis and P.~Agrawal, ``Walk these ways: Tuning robot control for
  generalization with multiplicity of behavior,'' \emph{arXiv preprint
  arXiv:2212.03238}, 2022.

\end{thebibliography}

\end{document}